\documentclass[runningheads]{llncs}
\usepackage{graphicx}
\usepackage{tikz}
\usepackage{comment}
\usepackage{amsmath,amssymb} 
\usepackage{color}
\usepackage{tabularx}
\usepackage{booktabs}
\usepackage[accsupp]{axessibility}
\usepackage{xspace}
\makeatletter
\DeclareRobustCommand\onedot{\futurelet\@let@token\@onedot}
\def\@onedot{\ifx\@let@token.\else.\null\fi\xspace}
\def\eg{\emph{e.g}\onedot}
\def\ie{\emph{i.e}\onedot}
\def\etal{\emph{et al}\onedot}
\makeatother
\newcommand{\minisection}[1]{\noindent{\textbf{#1}}}

\begin{document}
\pagestyle{headings}
\mainmatter

\title{Task Grouping for Multilingual Text Recognition} 
\titlerunning{Task Grouping for Multilingual Text Recognition}

\author{Jing Huang \and
Kevin J Liang \and
Rama Kovvuri \and
Tal Hassner}
\authorrunning{J. Huang et al.}
\institute{Meta AI \\
\email{\{jinghuang,kevinjliang,ramakovvuri,thassner\}@meta.com}}
\maketitle

\begin{abstract}
Most existing OCR methods focus on alphanumeric characters due to the popularity of English and numbers, as well as their corresponding datasets. On extending the characters to more languages, recent methods have shown that training different scripts with different recognition heads can greatly improve the end-to-end recognition accuracy compared to combining characters from all languages in the same recognition head. However, we postulate that similarities between some languages could allow sharing of model parameters and benefit from joint training. Determining language groupings, however, is not immediately obvious. To this end, we propose an automatic method for multilingual text recognition with a task grouping and assignment module using Gumbel-Softmax, introducing a task grouping loss and weighted recognition loss to allow for simultaneous training of the models and grouping modules. Experiments on MLT19 lend evidence to our hypothesis that there is a middle ground between combining every task together and separating every task that achieves a better configuration of task grouping/separation.

\end{abstract}

\section{Introduction}
\label{sec:intro}

Optical Character Recognition (OCR) has long been the fundamental task in computer vision. There are many applications such as automatic content extraction for documents \cite{burie2015icdar2015}, translation \cite{toyama2014mixed}, text style transfer \cite{krishnan2021textstylebrush} and assistance for robots and visually impaired users. From the perspective of research, OCR can range from relatively easy and controlled tasks such as digit recognition~\cite{lecun1998mnist}, to difficult scenarios such as scene text with arbitrary orientations and shapes \cite{karatzas2015icdar,ch2017total,hassner2013computation,chng2019icdar2019,singh2021textocr}, and has become an important domain for benchmarking new machine learning techniques.

Nevertheless, most existing OCR approaches focus on numbers and the English alphabet due to English's status as a common \textit{lingua franca} and its subsequent wide availability in popular datasets. Thanks to the introduction of multilingual text detection and recognition datasets and benchmarks \cite{nayef2017icdar2017,nayef2019icdar2019}, there is now a unified platform to measure the model performance on challenging scenarios containing thousands of distinctive characters. Recent methods have shown that training different languages with different recognition heads can improve end-to-end recognition accuracy compared to combining characters from all languages in the same recognition head \cite{huang2021multiplexed}. However, it's not clear whether an individual recognition head for each language is optimal. For example, should English and Spanish be separated into two heads? The answer is probably no since they share most of the characters. Moreover, even if separating two languages into two heads does yield the best accuracy, it might not be worth it if the accuracy gain is marginal compared to the increase in number of parameters and/or the inference time. Therefore, one of the questions our work tries to answer is how to decide whether/how languages should be grouped together under the constraint of a limited number of models.

Without any pre-assigned grouping, we treat each of the models at initialization as a generalist agent which looks at all tasks. Then, as the scale tips, each agent is encouraged to be increasingly specialized in one or more tasks; each agent becomes a specialist, each model can still try to learn the other tasks to some lesser extent. Due to different transferability, data variation, and similarities among the tasks, each agent can have different progress in both the specialized tasks and non-specialized tasks, and the specialties will be redistributed automatically as the agents evolve. Eventually, as confirmed by our experiments, this multi-agent system will reach an equilibrium where the specialties for each agent do not change any more, and this is when the task grouping result is finalized.

To summarize, our contributions include:
\begin{itemize}
\item To our knowledge, this is {\em the first work} exploring the grouping of languages for multilingual OCR.
\item We propose an automatic grouping mechanism that allows dynamic and differentiable routing of tasks to different heads during training.
\item We empirically show that the automatic task grouping model outperforms both the one-task-per-head and the all-tasks-in-one-head baselines.
We further show that when the models have different capacities, the task assignment can potentially reflect the underlying task complexity and data distribution.
\end{itemize}

To promote reproduction of our work, our code is publicly available at \url{https://github.com/facebookresearch/MultiplexedOCR}.

\section{Related work}

\subsection{Multilingual text spotting}
Text spotting systems combine text detection and text recognition modules to identify the location and content of text in images. Early works approached both modules independently: text proposals for regions containing text were generated first, followed by a recognition module to identify the text given a pre-defined character dictionary. For text detection, state-of-the-art (SotA) methods are mostly based on Region Proposal Networks (RPN)~\cite{ren2015faster}, previously proven successful for object detection. Variants of RPNs have been proposed to account for varying text orientations~\cite{jiang2017r2cnn}, arbitrary shapes~\cite{liao2020mask,qin2019towards}, and character masking~\cite{baek2019character}. For text recognition, models typically use an RNN-style design to predict sequences of characters~\cite{su2014str,he2016reading}. Representative methods include connectionist temporal classification (CTC)~\cite{graves2006connectionist} and attention-based decoders \cite{bahdanau2014neural,lee2016recursive,shi2016robust}. 

While earlier systems treated detection and recognition as independent modules, most of the recent works train these modules in an end-to-end manner. Given the inter-dependability of these modules, this training methodology results in performance improvements for these systems. Some representative works include Mask TextSpotter~\cite{liao2021mask}, FOTS~\cite{liu2018fots}, CharNet~\cite{xing2019charnet}, etc. For deeper insights into the text spotting systems, we refer the readers to the thorough review in \cite{long2020scene}. Similar to these works, we also employ end-to-end training for our system. For recognition, we mainly use an attention-based decoder to make fair comparisons with previous works~\cite{liao2020mask,huang2021multiplexed}. 

With the availability of reliable multilingual datasets such as MLT19~\cite{nayef2019icdar2019}, text spotting systems have tried to address the problem of multilingual texts. In addition to detection and recognition modules, some multilingual text spotting systems also include a script identification module~\cite{buvsta2018e2e,huang2021multiplexed} to identify the language for text recognition. While text spotting systems such as E2E-MLT~\cite{buvsta2018e2e} and CRAFTS~\cite{baek2020character} present results for multilingual datasets, they do not explicitly incorporate model specific components adapted for multiple languages. Instead, they combine the characters from all languages to form a larger dictionary for the recognition head. Recently, Multiplexed Multilingual Mask TextSpotter (MMMT) \cite{huang2021multiplexed} proposed to employ different recognition heads for different scripts, routed through a script identification module at the word level. Unlike MMMT, which employs hard assignment for routing to an appropriate recognition head, we propose to group the languages by training agents to route words to different heads in a data-driven fashion. This automatic grouping mechanism allows for dynamic and differentiable shifting of tasks to optimize the language combinations for various recognition heads.

\subsection{Multitask learning and grouping}

Multitask learning methods have a long history~\cite{caruana1997multitask,evgeniou2004regularized}. As their name implies, they jointly learn solutions for multiple tasks, sharing or transferring information between tasks to improve overall performance. Recent deep learning methods assume that the parameters for early layers, which account for low-level feature extraction, are shared among different tasks, while the parameters for later layers, which account for high-level integration of visual signals, are task-specific~\cite{zeiler2014visualizing,inkawhich2020transferable}. Hence, information relevant to all tasks is learned by a shared trunk which later splits to multiple task-specific heads. 

A natural question that arises when designing multitask systems is the following: How should tasks be grouped to maximise a model's accuracy? To answer this question, Kang \etal \cite{kang2011learning} proposed learning shared feature representations for related tasks by formulating task grouping as a mixed integer programming problem, where binary indicator variables are used to assign tasks to groups. Unlike this hard group assignment, Kumar \etal \cite{kumar2012learning} propose to allow for parameter sharing across groups through soft, latent assignment of task features as a linear combination of a finite number of underlying basis tasks. Zhong \etal \cite{zhong2016flexible} extend this work by removing constraints on the size of latent basis tasks and adding regularization terms to enforce sparsity in task weights and orthogonality to prohibit commonality among unrelated tasks. Zamir \etal \cite{zamir2018taskonomy} proposed a method for modeling the relationship of different visual tasks based on the {\em transferability} between them. Instead of learning shared representation on a trunk, Strezoski \etal \cite{strezoski2019routing} introduce a {\em Task Routing layer} that masks convolutional channels based on the task, effectively creating a sub-network per task.

Our work is similar in spirit to Strezoski \etal \cite{strezoski2019routing} in that we allow for dynamic routing of tasks to different heads during training. In our approach, however, the routing is done using {\em Gumbel-Softmax} \cite{jang2016categorical} to ensure probabilistic interpretation of each task and using a novel {\em grouping loss} for task assignment.

\begin{figure*}[t]
    \centering
    \includegraphics[width=0.98\linewidth]{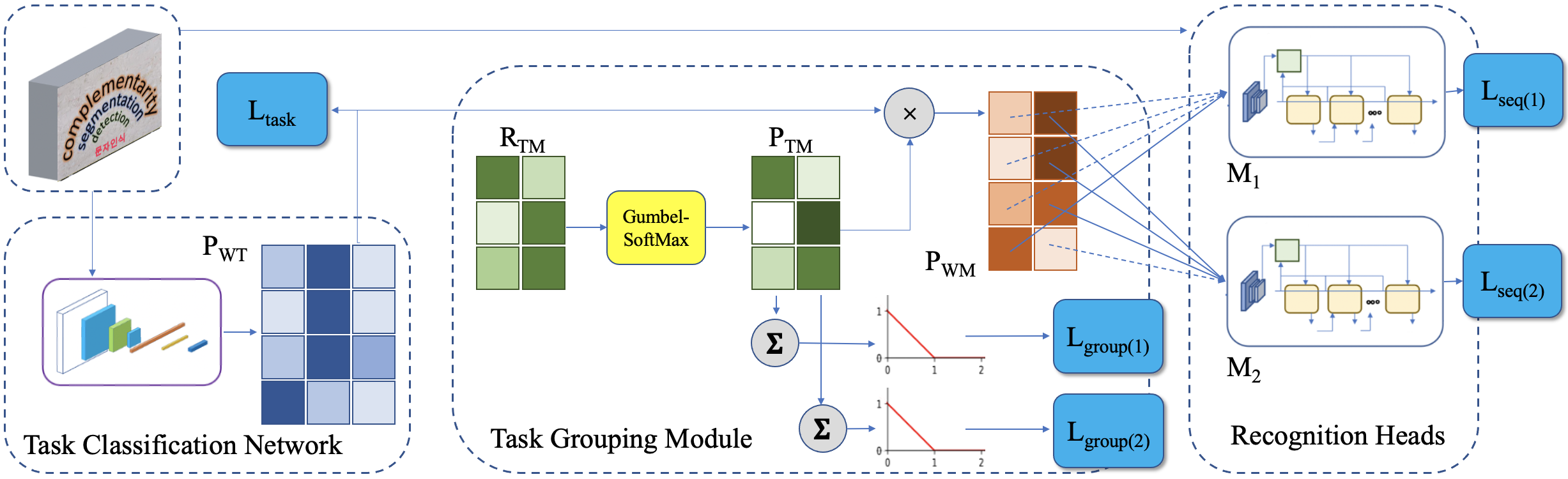}
    \caption{\textbf{Our proposed task grouping framework.} Here we show a batch of $4$ inputs potentially belonging to $3$ tasks, with $2$ recognition heads. See Sec.~\ref{sec:method} for more details.}
    \label{fig:task_grouper}
\end{figure*}

\section{Methodology}\label{sec:method}

Given a list of tasks $T = \{T_i\}_{1 \leq i \leq t}$ and a list of models $M = \{M_j\}_{1 \leq j \leq m}$, we can define the task grouping of $T$ over $M$ as a mapping $G: T \rightarrow M$, where $G$ is a single-valued function, which means each task will be assigned to exactly one model. On the other hand, $G$ does not need to be an injection, since multiple tasks can be assigned to the same model. $G$ need not be a surjection either, in which case some models will not be assigned with any tasks. Our goal is to find out the best assignment $G$ such that the overall performance is maximized.

Figure \ref{fig:task_grouper} shows the core architecture of the proposed task grouping framework. Given an input, which could be a batch of already cropped image patches or pre-extracted features, we first pass them through a {\em task classification network} that predicts the probabilities of each input belonging to each of the tasks.
Under the context of multilingual text recognition, each task $T_i$ can be described as ``recognizing the word instance $W_k$ in language set $L_i$'', where $W_k$ is the $k$-th instance in a batch of $w$ word crops. We can thus define a probability matrix of size $w \times t$ on the likelihood of each word belonging to each task/language set:

\begin{equation} \label{eq:word_task_matrix}
    P_{WT} = \{p(T_i | W_k)\}_{1 \leq k \leq w, 1 \leq i \leq t}
\end{equation}

At inference time, $P_{WT}$ can be inferred from a task classification network such as a language prediction network \cite{huang2021multiplexed}. This is a $t$-way classification problem, and the task classification loss can be computed using a cross entropy loss:
\begin{equation}
    L_{task}(W_k) = - \sum_{i=1}^{t} I(T_i=T_{gt}) \log p(T_i|W_k)
\end{equation}
where $I(T_i=T_{gt})$ is a binary indicator of the task matching the ground truth.

At training time, $P_{WT}$ can be inferred from the ground truth, if there is an annotation of which language each word belongs to: $p(T_i|W_k)$ is $1$ if $W_k$ belongs to $T_i$ and $0$ otherwise. When the ground truth annotation for the language information is not directly available but the transcription is available, we can make an educated guess of the probability by calculating the proportion of characters in $W_k$ that are supported by language set $L_i$.

\subsection{Grouping module}

Since task-model mapping $G$ is a discrete function, to be able to learn it we can define the following probability matrix, of size $t \times m$:
\begin{equation} \label{eq:task_model_matrix}
    P_{TM} = \{p(M_j|T_i)\}_{1 \leq i \leq t, 1 \leq j \leq m},
\end{equation}
where $p(M_j|T_i)$ is the probability of an arbitrary word belonging to $T_i$ to be handled by model $M_j$. Then, we can compute the probability matrix of each word $W_k$ to be handled by model $M_j$ by multiplying $P_{WT}$ and $P_{TM}$:

\begin{equation} \label{eq:word_model_matrix}
    P_{WM} = P_{WT} \cdot P_{TM}
\end{equation}
Naive task assignment to a group based on traditional SoftMax is a discrete operation and thus non-differentiable.
Backpropagation through only the selected task-group pairing would result in high variance gradients leading to unstable learning.
Instead, when computing $P_{WM}$ during training, we apply a soft relaxation of the assignment operation using the Gumbel-Softmax \cite{jang2016categorical}.
Gumbel-Softmax if fully differentiable with the reparameterization trick and results in gradient backpropagating through all possible task-group pairings, not just the one with the maximum score.
We instantiate learnable parameters for task-model assignment as a real-valued matrix $R_{TM} \in \mathbb{R}^{t \times m }$, initialized with all ones (or any equal numbers) in the beginning, and we set the temperature $\tau = 1.0$ throughout the training. At test time, we can just pick the model corresponding to the maximum, \ie the hard mode of Gumbel-Softmax.

\subsection{Integrated loss}

A key difference of our approach compared to \cite{huang2021multiplexed} is that in our framework, we do not restrict the capability of each model, or recognition head, to support any specific task, \ie, a certain recognition head can only support certain characters, from the beginning. Instead, we assume each model to be omnipotent in the beginning and has the potential to handle every task. This is necessary since otherwise there is no point in doing the grouping if each model is already designed to do certain tasks. 

Therefore, unlike \cite{huang2021multiplexed}, we can directly use the negative log likelihood as the recognition loss $L_{seq(j)}$ for each model $M_j$ without worrying about the unsupported characters:
\begin{equation}
    L_{seq} = - \frac{1}{s} \sum_{l=1}^{s} \log p(S_l),
\end{equation}
where $p(S_l)$ is the predicted probability of character at position $l$ of the sequence, and $s$ is the length of the sequence of character labels.
We can, however, perform the pruning at the output layer to remove any characters that do not belong to the task assigned to certain head, once the grouping is determined. This reduces the unnecessary weights in the final model.

The integrated loss across all probabilistic instance-model assignments can thus be calculated as the weighted sum of individual losses:

\begin{equation}
    L_{integrated0} = \sum_{k=1}^{w}\sum_{j=1}^{m} p(M_j|W_k) \cdot L_{seq(j)}(W_k, M_j),
    \label{eq:integrated_loss_0}
\end{equation}
where the probability term is from $P_{WM}$ of Eq. \eqref{eq:word_model_matrix}, which is essentially the law of total probability:

\begin{equation}
    p(M_j|W_k) = \sum_{i=1}^{t} p(T_i|W_k) \cdot p(M_j|T_i)
    \label{eq:word_model_element}
\end{equation}

\subsection{Integrated loss with a base loss coefficient}

With the integrated loss (Eq. \eqref{eq:integrated_loss_0}), we can see that in general, a task $T_{big}$ with a bigger probability $p(M_j|T_{big})$ to be assigned to a model $M_j$ will contribute a bigger loss than a task $T_{small}$ with a smaller probability $p(M_j|T_{small})$ to be assigned to the model, encouraging the model to optimize towards a better prediction for $T_{big}$, which then encourages $p(M_j|T_{big})$ to be bigger until it reaches $1$. A similar but opposite process applies to $p(M_j|T_{small})$, which would become smaller until it reaches $0$. As a result, the learned task-model assignment $P_{TM}$ will almost certainly be random and fully depending on the first few iterations due to the positive-feedback loop. We resolve this issue by adding a small positive base loss coefficient, $\epsilon$:

\begin{equation}
    L_{integrated} = \sum_{k=1}^{w}\sum_{j=1}^{m} (p(M_j|W_k)+\epsilon) \cdot L_{seq(j)}(W_k, M_j).
    \label{eq:integrated_loss}
\end{equation}

This ensures that the model not only tries to excel at the tasks assigned to it, but also learns the other tasks at a small but positive rate. The effect of $\epsilon$ can be quantified from the perspective of training data ratios among different tasks. Assume the original ratio of data from any task is $1$, for any model-task pair, the maximum effective data ratio would be $1 + \epsilon$, which is achieved when $p$ reaches $1$, and the minimum effective data ratio would be $0 + \epsilon$, which is achieved when $p$ falls to $0$. The ratio $\frac{1 + \epsilon}{\epsilon}$ can thus be used to measure how biased the model can potentially be trained towards the most vs. least important task. Based on our ablation study (Sec. \ref{sec:ablation_study}), we set $\epsilon=0.2$ when training from scratch, $\epsilon=0.1$ when fine-tuning from pretrained models and $\epsilon=0$ for the final head-wise fine-tuning.

\subsection{Grouping loss}

While Eq. \eqref{eq:integrated_loss} makes sure that any model has the potential to learn every task, we also would like to ensure that happens within a certain budget, i.e., given the number of different models (heads) we can support, each model is specialized in at least one task. This ensures we do not waste the modeling capacity of an idle head. Therefore, we introduce the following grouping loss

\begin{equation}
    L_{group} = \sum_{j=1}^{m} L_{group(j)} \\
    = \sum_{j=1}^{m}\max(\mu_j - \sum_{i=1}^{t}p(M_j|T_i), 0),
    \label{eq:grouping_loss}
\end{equation}
where $\mu_j$ is the least number of tasks model $M_j$ is expected to handle. In most experiments, we set $\mu_j=1$, meaning that if $M_j$ completely takes over at least one task, the grouping loss for $M_j$ would reach the minimum value $0$. Note that the converse does not hold - the grouping loss can reach $0$ even when certain model do not excel in any specific task. However, in practice, as long as the number of tasks is larger than or equal to the number of models, the small penalty of the grouping loss could help us achieve the minimum task assignment goal.

\section{Experimentals}

\subsection{Datasets}
Our work leverages a number of public datasets. These sets are summarized in Table~\ref{tab:datasets}. We next offer a brief description of these sets.

\minisection{ICDAR 2013 dataset (IC13)}~\cite{karatzas2013icdar} This is the oldest set used in this work, originally released for the ICDAR 2013 Robust Reading Competition. It offers 229 training and 233 test images of English text. Text locations are given as axis aligned, rectangular bounding boxes with text annotated at a word level. 

\minisection{ICDAR 2015 dataset (IC15)}~\cite{karatzas2015icdar}
This dataset was introduced in ICDAR'15 and offers more images than IC13: 1000 training and 500 test. Images in this set are of scene text in English, appearing at different orientations, where words are annotated using quadrangle bounding boxes.

\minisection{Total Text dataset}~\cite{ch2017total} This collection offers 1255 training and 300 test, English scene text images. The images reflect a wide range of text orientations and shapes, including curved text examples. To accommodate different shapes, text locations are provided as polygons; recognition labels are given at word level.

\minisection{ICDAR 2017 RCTW dataset (RCTW17)}~\cite{shi2017rctw} This set was collected to promote development of OCR methods for in the wild Chinese text. It is partitioned to 8034 and 4229 subsets of training and test images, respectively.

\minisection{ICDAR 2019 MLT dataset (MLT19) and SynthTextMLT}~\cite{nayef2019icdar2019} was an extension of the ICDAR 2017 MLT dataset (MLT17) \cite{nayef2017icdar2017} for multilingual text detection, recognition and script identification, which contains 10000 training images, 2000 validation images and 10000 test images in 7 different scripts from 10 languages. The dataset contains multi-oriented scene text annotated by quadrangle boxes. A synthetic dataset (SynthTextMLT)~\cite{buvsta2018e2e} containing over 250k synthetic data in 7 scripts was also released along with the MLT19 benchmark. Since MLT19 training and validation sets completely covers the training and validation images in MLT17, though the split is a bit different, we only use MLT19 data for training in this paper.

\minisection{ICDAR 2019 ArT dataset (ArT19)}~\cite{chng2019icdar2019} Contains 5603 training and 4563 test images in both English and Chinese, sourced from Total Text~\cite{ch2017total} and SCUT-CTW1500~\cite{yuliang2017detecting}. Released as part of the ICDAR 2019 Robust Reading Competition, the images in this collection depict texts in challenging shapes. Similarly to Total Text, text locations are encoded as polygons. We remove all Total Text test images from this set, ensuring that any training on this set can be applied to other sets without risk of test images influencing models trained on this set.

\minisection{ICDAR 2019 LSVT dataset (LSVT19)}~\cite{sun2019lsvt}
This is one of the largest data sets used for developing OCR methods: 30000 training and 20000 test images. LSVT images mostly show street views with about 80\% of them showing Chinese text and the rest examples in English. 

\begin{table}[t]
    \scriptsize
    \centering
    \caption{\textbf{Datasets used in our experiments.} \#Train: number of training images. Ratio: the relative sampling ratio when the dataset is used in training. Word / Phrase: Annotations given at a word or phrase level. Box type: horizontal, axis aligned (H-Box), arbitrarily rotated (R-Box), quadrangle (Quad), and Polygon. \#Lang: Number of languages provided. Note that the Total Text dataset is fully covered in ArT19, and we removed the testing set of Total Text from ArT19.}
    \resizebox{0.95\columnwidth}{!}{
    \begin{tabular}{lccccc}
    \toprule
    Name & \#Train & Ratio & Word / Phrase & Box type & \#Lang.\\
    \midrule
    ICDAR13~\cite{karatzas2013icdar}         & 229 & 20 & Word & H-Box & 1 \\  
    ICDAR15~\cite{karatzas2015icdar}         & 1000 & 20 & Word & Quad & 1 \\
    Total Text~\cite{ch2017total} & 1255 & 50 & Word & Polygon & 1 \\
RCTW17~\cite{shi2017rctw}          & 8034 & 20 & Phrase & R-Box & 2 \\    
MLT19~\cite{nayef2019icdar2019}           & 10000 & 100 & Word & Quad & 10 \\
SynthTextMLT~\cite{buvsta2018e2e}    & 252599 & 1 & Word & R-Box & 7 \\
ArT19~\cite{chng2019icdar2019}           & 5303 & 50 & Word & Polygon & 2 \\
LSVT19~\cite{sun2019lsvt}          & 30000	& 20 & Phrase & Polygon & 2 \\
    \bottomrule
    \end{tabular}
    }
    \label{tab:datasets}
\end{table}

\subsection{Model training}
We base our implementation on the Multiplexer codebase\footnote{\url{https://github.com/facebookresearch/MultiplexedOCR}}.
For fair comparison, we adopt the same segmentation-based detection and ROI mask feature extraction modules as \cite{huang2021multiplexed}, and freeze the pretrained weights of these layers throughout training. For language classification, \cite{huang2021multiplexed} uses 8 classes including Arabic, Bengali, Chinese, Hindi, Japanese, Korean, Latin, and Symbol, but in our experiment we only use 7 classes by discarding the Symbol class, as it does not have any dedicated dataset and the number of the samples is too small to make a difference.

To expedite the training, we first combine every dataset to train a single recognition head with hidden size 256 and embed size of 200 covering all datasets using the ratios specified in Table \ref{tab:datasets} for 40k iterations. We then use this weight as a universal pretrained weights for the second stage of training.

Next, we perform a series of experiments that jointly train the grouping module and the recognition heads, each restricting the number of recognition heads to $m$ ($2 \leq m \leq 7$). For each $m$, we launch three training jobs with different random seeds. Each of the training jobs runs for 20k iterations on the MLT19 training datasets only to reduce the potential data imbalance when including the other training set. We record and summarize the final grouping results in Table \ref{tab:task_grouping_result}, which we will discuss in \ref{sec:results:grouping}.

Finally, based on the grouping result, we fine-tune each recognition head with only the datasets within the assigned group corresponding to the head. At this stage the grouping is essentially frozen and does not change any more. We can prune the output layer of the decoder so that the characters not belonging to the group are removed, to reduce the parameter number for the final model.

\subsection{Task grouping results}
\label{sec:results:grouping}

Table \ref{tab:task_grouping_result} shows the aggregation of grouping results from $18$ task grouping experiments with $2$ to $7$ recognition heads, each repeated for $3$ times. All task assignments stabilize after about 10000 iterations.

The top 14 groups are ordered first by the number of occurrences and then by the first occurrence, \ie the minimum number of recognition heads when the group first occurs. All exclusive task-model assignments (one head focusing on one task) occur at least twice, showing the effectiveness of having a dedicated model for each task. Chinese ending up as an individual task occurs in 50\% of the cases, which is expected given its high character number and the datasets, except that it's grouped together with Japanese, which shares many characters with it, only once. On the other hand, Hindi seems to be suitable to be grouped with many different languages rather than being trained by itself. 

Surprisingly, the most frequent task group that has more than one task is Arabic+Korean, which occurs 5 times. This suggests that there are inherent characteristics shared either by these two scripts, or by the examples in the MLT19 dataset itself, that boost the performance for each other. Another unusual cluster is the combination of 5 tasks, Bengali+Chinese+Hindi+Japanese+Latin, which is the only grouping with more than 2 tasks that occurs more than once.
We note however that the scattering of the grouping results shows that there can be many local optima for this specific scenario of $7$ distinctive scripts. We shall expect higher frequencies of the same grouping results if certain tasks share greater similarity, and we will leave that as one of our future work.
We additionally find it interesting that despite a grouping loss to encourage each head to take on at least one task, we observe that some recognition heads might not be assigned with any task in the end when there are $6$ or $7$ tasks. This means for certain combinations of tasks, training them together could outperform training them separately, even if there is spare resource for a new head.

\begin{table}[t]
    \scriptsize
    \centering
    \caption{\textbf{Task grouping result.} Task combinations that end up being grouped together. The 2nd to the 5th columns indicate task names in the final grouping, the number of tasks in the group, the number of occurrences in the 18 experiments, and the minimum number of recognition heads when the combination first occurs. }
    \begin{tabularx}{0.98\linewidth}{c|X|c|c|c}
    \toprule
    Rank & Group & \#Tasks within group & \#Occurrences & \#Heads at first occurrence \\ 
    \midrule
    1 & Chinese (C) & 1 & 9 & 4 \\
    2 & Latin (L) & 1 & 7 & 5 \\
    3 & Arabic (A) & 1 & 6 & 3 \\
    3 & Korean (K) & 1 & 6 & 3 \\
    5 & Arabic+Korean & 2 & 5 & 2 \\
    6 & Bengali (B) & 1 & 5 & 5 \\
    7 & Japanese (J) & 1 & 4 & 5 \\
    8 & B+C+H+J+L & 5 & 2 & 2 \\
    9 & Hindi+Japanese & 2 & 3 & 3 \\
    10 & Japanese+Latin & 2 & 2 & 4 \\
    11 & Arabic+Hindi & 2 & 2 & 5 \\
    11 & Bengali+Japanese & 2 & 2 & 5 \\
    11 & Hindi+Latin & 2 & 2 & 5 \\
    14 & Hindi (H) & 1 & 2 & 6 \\
    \hline
    15 & A+K+L & 3 & 1 & 2 \\
    15 & H+J+K & 3 & 1 & 2 \\
    15 & A+B+C+L & 4 & 1 & 2 \\
    15 & B+C+H+J & 4 & 1 & 2 \\
    15 & Chinese+Hindi & 2 & 1 & 3 \\
    15 & Korean+Latin & 2 & 1 & 3 \\
    15 & A+B+J & 3 & 1 & 3 \\
    15 & B+C+L & 3 & 1 & 3 \\
    15 & Arabic+Bengali & 2 & 1 & 4 \\
    15 & Bengali+Hindi & 2 & 1 & 4 \\
    15 & Chinese+Latin & 2 & 1 & 4 \\
    15 & A+B+H & 3 & 1 & 4 \\
    15 & Japanese+Korean & 2 & 1 & 5 \\
    15 & B+C+K & 3 & 1 & 6 \\
    15 & Hindi+Korean & 2 & 1 & 7 \\
    15 & Chinese+Japanese & 2 & 1 & 7 \\
    \bottomrule
    \end{tabularx}
    \label{tab:task_grouping_result}
\end{table}

\subsection{Ablation study}
\label{sec:ablation_study}

\minisection{Base integrated loss coefficient.} We train the task grouping network with different base integrated loss coefficient $\epsilon$ defined in Eq. \eqref{eq:integrated_loss} on MLT19 training set. The network contains 5 recognition heads that are initialized with the same pretrained weights. We record the number of task assignment changes in the first 3000 iterations. From Table \ref{tab:base_loss_weight} we can see that, when $\epsilon=0.0$, there's only 1 assignment change since the model does not have much chance to learn the unassigned tasks; interestingly, when $\epsilon$ is too big ($0.3/0.4$), there are also fewer changes happening, possibly because there is not much diversity across the models and everything moves in the same direction. The maximum number of assignment changes happen when $\epsilon$ is $0.2$ or $0.1$. Therefore, in most of our experiments we use $0.2$ for early training and $0.1$ for fine-tuning.

\begin{figure*}[ht]
    \centering
    \includegraphics[width=0.98\linewidth]{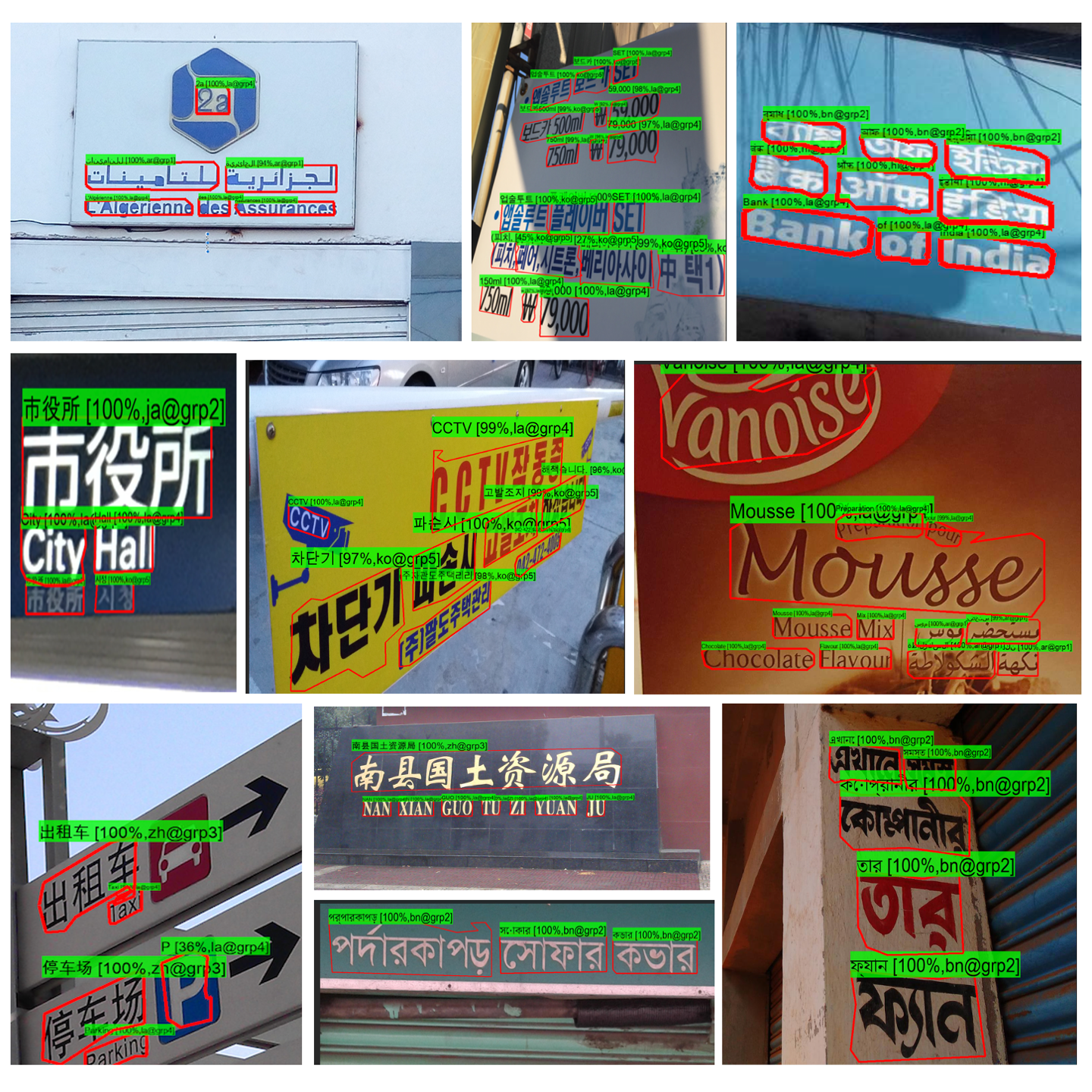}
    \caption{\textbf{Qualitative results on MLT19 test set~\cite{nayef2019icdar2019}.} The predicted transcription is rendered with green background, along with the detection confidence, language and the assigned group. The model has $5$ heads (groups): group 1 - Arabic (ar) and Hindi (hi), group 2 - Bengali (bn) and Japanese (ja), group 3 - Chinese (zh), group 4 - Latin (la), group 5 - Korean (ko). See Sec.~\ref{sec:results:e2e} for more details.}
    \label{fig:qualitative}
\end{figure*}

\begin{table}[t]
    \scriptsize
    \centering
    \caption{\textbf{Ablation study for base integrated loss coefficient $\epsilon$.} }
    \begin{tabularx}{0.98\linewidth}{X*{5}{p{0.1\linewidth}}}
    \toprule
    Base integrated loss coefficient $\epsilon$ & 0.0 & 0.1 & 0.2 & 0.3 & 0.4 \\ 
    \midrule
    Assignment changes within 3k iters & 1 & 3 & 5 & 2 & 2 \\
    \bottomrule
    \end{tabularx}
    \label{tab:base_loss_weight}
\end{table}

\begin{figure}[ht]
    \centering
    \includegraphics[width=0.9\linewidth, scale=0.8,clip,trim=0mm 2mm 0mm 2mm]{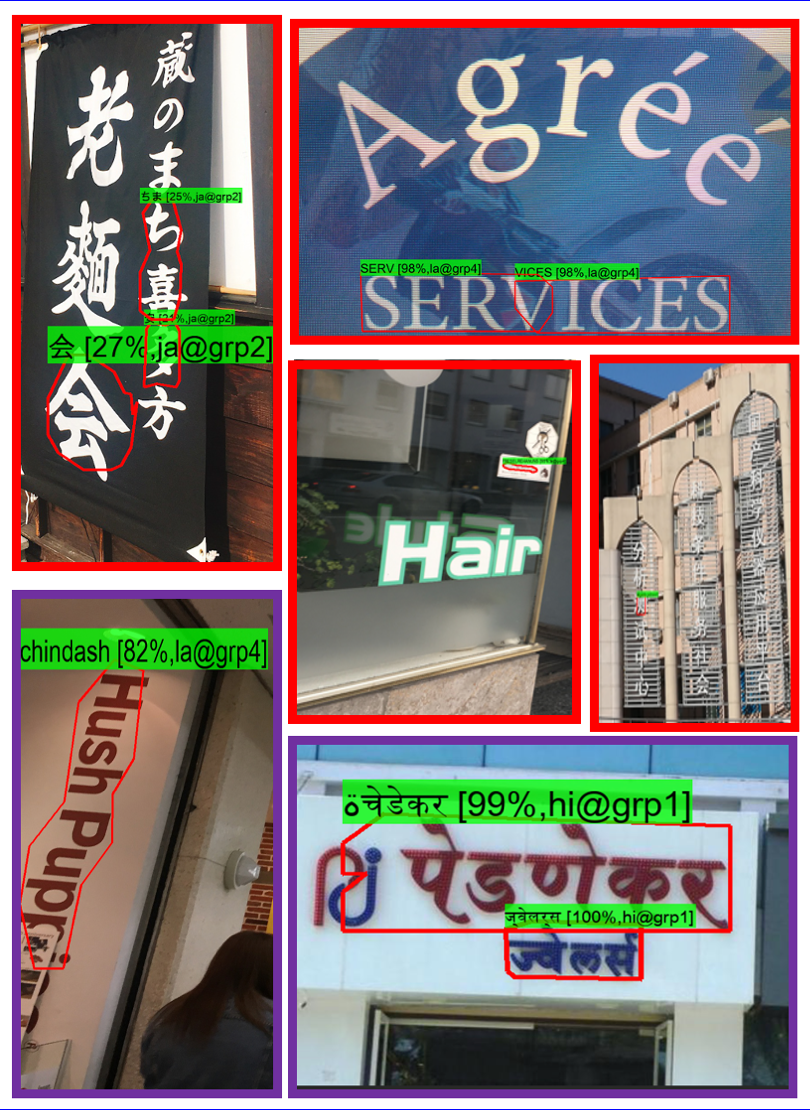}
    \caption{\textbf{Error analysis on MLT19.} Detection errors are represented in red outline and recognition errors in purple. See Sec.~\ref{sec:results:e2e}.}
    \label{fig:failures}
\end{figure}

\subsection{Task assignment on models with different hyper-parameters}

In this section, we perform an interesting experiment that showcases how our design can help assign different tasks to models with different hyper-parameters based on the potential difficulty and the available data. We set the number of models (recognition heads) to be equal to the number of tasks, but set the key hyper-parameters of the models, embed size and hidden size, to be different from each other. We train the overall model on the weighted combination of all datasets listed above, and Table \ref{tab:parameter_number_comparison} shows the assigned task corresponding to each of the models. We can clearly see the correlation between the number of parameters versus the number of characters in the corresponding character set, with the exception of Latin. This illustrated that in general, when the number of characters grow, the heavier models will outperform lighter models in the long term; however, since Latin words are dominating in all the datasets including many difficult cases like curved text, the task grouping framework learns to spend the heaviest model on it to boost the overall performance.

\begin{table}[t]
    \scriptsize
    \centering
    \caption{\textbf{Task assignment result for models with different major hyper-parameters.} Each model supports all characters in the beginning so the total number of parameters for each head is high, but they can be pruned when the task assignment stabilizes.}
    \begin{tabularx}{0.98\linewidth}{c|c|c|c|c|c}
    \toprule
    Embed Size & Hidden Size & Parameter Number & Assigned Task & Charset Size & Final Parameters\\
    \hline
    100 & 224 & 4.05M & Arabic & 80 & 1.15M \\
    150 & 224 & 4.51M & Bengali & 110 & 1.18M \\
    200 & 224 & 4.98M & Japanese & 2300 & 2.13M \\
    100 & 256 & 4.59M & Hindi & 110 & 1.42M \\
    150 & 256 & 5.06M & Korean & 1500 & 2.00M \\
    200 & 256 & 5.52M & Chinese & 5200 & 3.78M \\
    250 & 256 & 5.98M & Latin & 250 & 1.54M \\
    \bottomrule
    \end{tabularx}
    \label{tab:parameter_number_comparison}
\end{table}

\subsection{E2E text recognition}\label{sec:results:e2e}

Table \ref{tab:mlt19_task4} shows the results on MLT19 \cite{nayef2019icdar2019} end-to-end multilingual recognition benchmark. Fig.~\ref{fig:qualitative} additionally provides qualitative examples of these multilingual. We find that using varying numbers of grouped heads can perform similarly to (and in some cases, better than) the multiplexed approach of a separate recognition head per language~\cite{huang2021multiplexed}.
This is an interesting result, as it means we can significantly cut down on the computational cost and model size with little impact or even some gains to the performance.
Notably, we also find that increasing the number of heads from a single shared head (Mask TextSpotter V3 \cite{liao2020mask}) to even just two grouped heads leads to a significant increase in F1-score. 

We provide qualitative failure cases in Fig.~\ref{fig:failures}. While detection errors could be attributed to arbitrary text shape, blurred text, glossy surfaces and rare fonts; recognition errors could be attributed to text ordering, text resolution and challenging scripts. 

\begin{table}[t]
    \scriptsize
    \centering
    \caption{\textbf{End-to-end recognition results on MLT19.} Note that there are two versions results of CRAFTS, one from the official MLT19 website and one from paper \cite{baek2020character}. Importantly, CRAFTS has a ResNet-based feature extraction which is much bigger than the one with $5$-Convs used in our experiments.}
    \begin{tabularx}{0.9\linewidth}{lX*{2}X}
    \toprule
    Method & F & P & R \\ 
    \midrule
    E2E-MLT \cite{buvsta2018e2e} & 26.5 & 37.4 & 20.5 \\
    RRPN+CLTDR \cite{ma2018arbitrary} & 33.8 & 38.6 & 30.1 \\
    CRAFTS \cite{baek2020character} & 51.7 & 65.7 & 42.7 \\
    CRAFTS (paper) \cite{baek2020character} & \textbf{58.2} & \textbf{72.9} & \textbf{48.5} \\
    \hline
    Mask TextSpotter V3 (1 head) \cite{liao2020mask} & 39.7 & \textbf{71.8} & 27.4 \\ 
    Multiplexed TextSpotter (8 heads) \cite{huang2021multiplexed} & 48.2 & 68.0 & 37.3   \\
    \hline
    Grouped (2 heads) & 45.5 & 67.7 & 34.3 \\
    Grouped (3 heads) & 47.1 & 67.0 & 36.3 \\
    Grouped (4 heads) & 47.9 & 66.7 & 37.4 \\
    Grouped (5 heads) & \textbf{48.5} & 67.7 & \textbf{37.8} \\
    Grouped (6 heads) & 48.3 & 67.8 & 37.5 \\
    Grouped (7 heads) & 48.2 & 68.0 & 37.3 \\
    \bottomrule
    \end{tabularx}
    \label{tab:mlt19_task4}
\end{table}

\section{Conclusions}
Text is one of the most ubiquitous visual object classes in real-world scenes, making understanding it practical and critically important. Processing multiple languages, however, requires substantial resources, to accurately recognize the subtleties of appearances variations of different scripts and different intra-script characters. This ability was, therefore, previously accomplished by specializing separate network heads to specific languages. We, instead, are the first to propose automatically grouping different languages together, in the same recognition heads. Our dynamic, and differentiable task shifting approach automatically routes tasks to different heads while the network trains, optimizing for the best, bottom line accuracy across all languages. Extensive tests show our method to not only achieve SotA accuracy, but to do so with fewer recognition heads and hyperparameters, consequently making is a practical design choice for real-world OCR systems. 

\minisection{Future work} Our work leaves several natural follow-up directions. One interesting question relates to the scalability of our approach: How many multitask heads, for example, would be required to effectively learn hundreds of languages? Another intriguing direction is extending our multitask learning and task grouping to include neural architecture search as part of its design. Such a solution should allow growing heads with different architectures for different languages to account for, \eg, harder vs. easier languages. Finally, another potential extension could be continual language learning~\cite{Parisi2019}: adding more languages as relevant training data becomes available, without retraining or regrouping existing languages. Alternative grouping approaches based on Bayesian nonparametric approaches like the Chinese Restaurant Process~\cite{Aldous1985} or Indian Buffet Process~\cite{Ghahramani2006,mehta2021continual} may be natural ways to perform groupings in such settings.

\clearpage
\bibliographystyle{splncs04}
\bibliography{bib}
\end{document}